\documentclass{article}

     \usepackage[preprint]{neurips_2019}

\usepackage{amsmath}
\usepackage{amssymb}
\usepackage{amsthm}
\usepackage{graphicx}
\usepackage{graphics}
\usepackage{subcaption}
\usepackage{wrapfig}

\usepackage[utf8]{inputenc} % allow utf-8 input
\usepackage[T1]{fontenc}    % use 8-bit T1 fonts
\usepackage{hyperref}       % hyperlinks
\usepackage{url}            % simple URL typesetting
\usepackage{booktabs}       % professional-quality tables
\usepackage{amsfonts}       % blackboard math symbols
\usepackage{nicefrac}       % compact symbols for 1/2, etc.
\usepackage{microtype}      % microtypography

\title{Detailed comparison of communication efficiency of split learning and federated learning}

\author{%
  Abhishek Singh, Praneeth Vepakomma, Otkrist Gupta, Ramesh Raskar \\
  Massachusetts Institute of Technology\\
  Cambridge, MA 02139 \\
  \texttt{vepakom@mit.edu} \\
}

\begin{document}

\maketitle

\begin{abstract}
 
 We compare communication efficiencies of two compelling distributed machine learning approaches of split learning and federated learning. We show useful settings under which each method outperforms the other in terms of communication efficiency. We consider various practical scenarios of distributed learning setup and juxtapose the two methods under various real-life scenarios. We consider settings of small and large number of clients as well as small models (1M - 6M parameters), large models (10M - 200M parameters) and very large models (1 Billion-100 Billion parameters). We show that increasing number of clients or increasing model size favors split learning setup over the federated while increasing the number of data samples while keeping the number of clients or model size low makes federated learning more communication efficient.  

\end{abstract}
\section{Introduction}
% Motivation for communication efficiency in the distributed learning setup.
Recent advances in deep learning has enabled ubiquitous applications in society and rapid growth of devices, data has ushered the need for distributed machine learning. Federated learning~\citep{federated_1,konevcny2016federated} and Split learning~\citep{split_1,split_2,vepakomma2019reducing} are two methods which allow to train a model collectively from various distributed data sources without sharing raw data. However, with such increasing number of devices (data sources) and increasing model complexity it is important to understand the role of these factors on communication efficiency of these distributed learning algorithms. \begin{wrapfigure}{r}{0.25\textwidth\label{splitfig1}
  \begin{center}
   \includegraphics[width=0.19\textwidth]{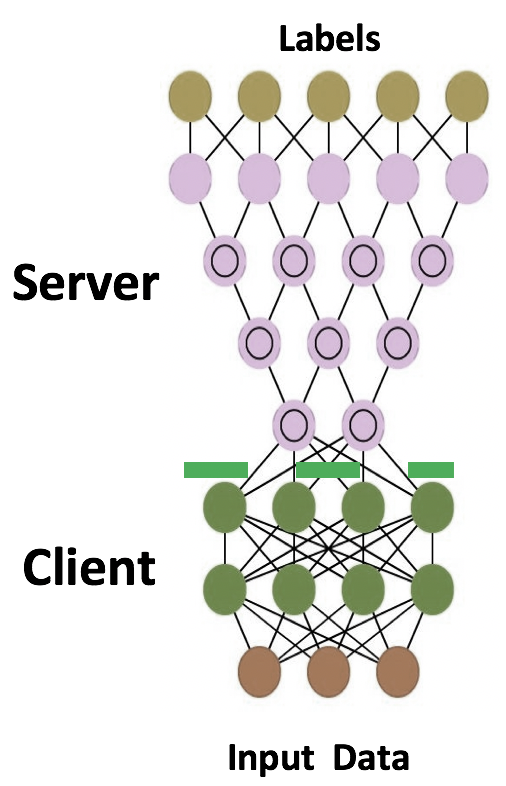}}
    \caption{Vanilla split learning setup showing distribution of layers across client and server.}
  \end{center}
  
\end{wrapfigure} In this work, we compare the communication efficiency of federated learning and split learning that allow training of deep neural networks from multiple data sources in a distributed fashion while not sharing the raw data in data sensitive applications.

Let us for example consider a network of data sources such as smart watches, hospitals, word corpus models or biobanks. 
Each of these entities have varying amounts of labeled data which we would like to use to train a deep learning pipeline. We ask ourselves on how one can train in a distributed setting without using too much communication bandwidth or computational burden at each of these locations? \par We now describe split learning, federated learning in sections 1.1 and 1.2 and then compare in detail the communication efficiencies of both these approached in section 2 followed by further analysis of these efficiencies with various dataset sizes, increasing number of clients and model sizes in section 3.

\subsection{Split learning} The method for training split learning has two main parts, a.) the topology step and b.) the training step. The topology step involves dividing the neural network into two separate parts comprised of some layers in the beginning and remaining layers at the end as shown in Figure  \ref{splitfig1}
.\begin{wrapfigure}{r}{0.37\textwidth}\label{split2}
  \begin{center}
    
   \includegraphics[width=\linewidth]{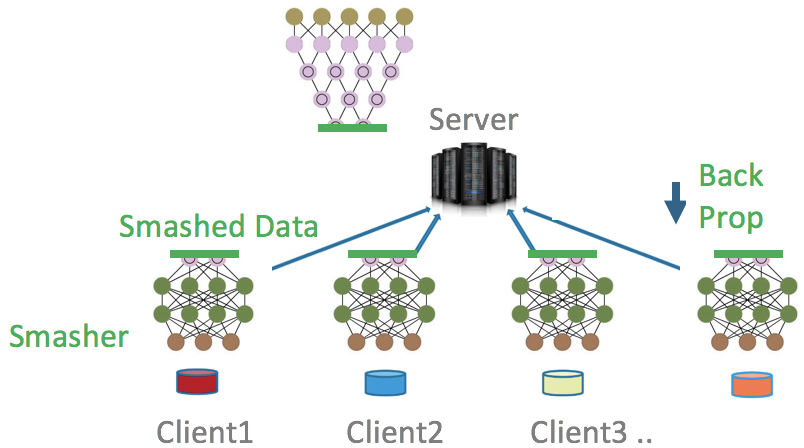}

  \end{center}
  \caption{Split learning setup with multiple clients and a server with dotted green line showing the split between the client's share of layers and the server's share of layers. Activations from only the split layer (last layer of client) are shared during forward propagation and gradients from only first layer of server are shared with client during backpropagation.}
\end{wrapfigure}
Both sides initialize network randomly and proceed to the following training steps. 

The training steps involve forward propagation of data through the beginning layers by data source. In the simplest of configurations, the output tensor from these layers and the corresponding label is then transmitted over to the cloud. The cloud continues the forward propagation by processing the output tensor through its remaining layers. The cloud then computes gradients using the transmitted label and propagates the gradient backward. The gradient generated at the first layer of server is then transmitted back to client (data source) and these steps are repeated until convergence. By following these steps we can actually train the deep network without requiring sharing of raw data by client, or any details of the part of the model held by client or server. More advanced configurations of split learning where there is no label sharing or configurations for multi-task learning with vertically partitioned data, or multi-hop 'Tor' like communication are detailed in \cite{split_2}. In the case of multiple clients and one server, there are two approaches of training, one with weight synhcronization between any client $i$ and the next client $i+1$ after client $i$ completes an epoch or across batches, and the other approach is without any client weight synchronization where clients take turns with alternating epochs in working with the server.
\subsection{Federated learning} In this other approach for distributed learning, every client runs a copy of the entire model on its own data. The server receives the weight updates from every client and averages them to get the updated weights from the server. 
\begin{wrapfigure}{r}{0.3\textwidth}\label{splitTheo}
  \begin{center}
   \includegraphics[width=0.25\textwidth]{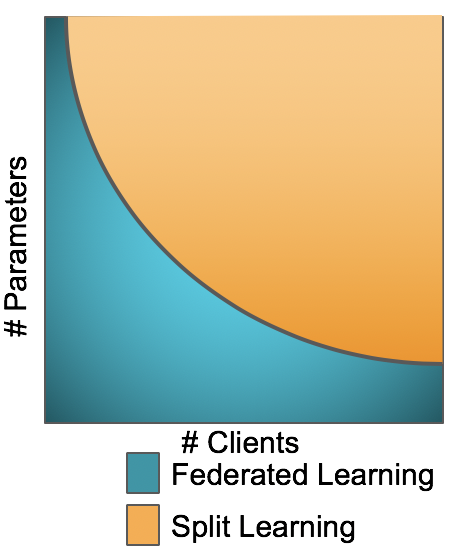}
  \end{center}
  \caption{Hyperbola dividing the regions where one technique perfors better over the other and both the feasible regions are shown in this theoretical schematic cartoon. Real instances of this equation are shown in the Analysis section.}
  \label{Split-learning}
\end{wrapfigure}
The updated weights are then downloaded by the clients and the process continues until convergence. 

\section{Communication efficiency}
In this section we describe our calculations of the communication efficiency for both of the distributed learning setups of split learning and federated learning. For analyzing the communication efficiency, we consider the amount of data transferred by every client for the training and client weight synchronization since rest of the factors affecting the communication rate is dependent on the setup of training cluster and is independent of the distributed learning setup. We use the following notation to mathematically measure the communication efficiencies, \par
\textbf{Notation:}  $K = $ \# clients, $N =$ \# model parameters, $p =$ total dataset size, $q =$ size of the smashed layer, $\eta$ = fraction of model parameters (weights) with client and therefore $1-\eta$ is fraction of parameters with server. \par In Table 1 we show the communication required per client per one epoch as well as total communication required across all clients per one epoch. As there are $K$ clients, when size of the training dataset across each client is the same, there would be $p/K$ data records per client in split learning. Therefore during forward propagation the size of the activations that are communicated per client in split learning is $(p/K)q$ and during backward propagation the size of gradients communicated per client is also $(p/K)q$. In the case where there is client weight sharing, passing on the weights to next client would involve a communication of $\eta N$.
\par In federated learning the communication of weights/gradients during upload of individual client weights and download of averaged weights are both of size $N$ each. 
\begin{table}[!htbp]
\label{formula_table}
\begin{tabular}{|l|l|l|}
\hline
Method                                       & Communication per client       & Total communication \\ \hline
Split learning with client weight sharing    & $(p/K)q + (p/K)q + \eta N$ & $2pq+ \eta NK$  \\ \hline
Split learning with no client weight sharing & $(p/K)q + (p/K)q$          & $2pq$           \\ \hline
Federated learning                           & $2N$                       & $2KN$           \\ \hline
\end{tabular}
\caption{Communication per client and total communication for the distributed learning setup as measured by the data transferred by all of the nodes in the learning setup.}
\end{table}
The split learning setup has two variants where one involves weight sharing among clients while the other variant does not. Sharing weights among clients helps in the synchronization among the clients but at the same time leaks more information as held by the weights of the model. Weight sharing among client adds extra communication overhead where the amount of overhead depends upon the model size ($N$) and the size of the smashed layer ($q$). The variant that does not involve weight sharing is based on alternating turns of epochs taken by the clients in working with the server.
The communication efficiency computed as ratio of data transfers of federated learning and split learning therefore is given by 
\label{comm_eff}
$
    \rho = \frac{2NK}{2pq+\eta NK}
$.
Split learning wins in terems of communication efficiency in scenarios when $\rho > 1$ and federated learning wins when $\rho < 1$.  Upon rearranging the terms, and expressing it as an equality, and dividing the numerator and denominator by $NK$ we get the equation of a rectangular hyperbola as $N=\frac{2pq}{((2-\eta) K)}$ in the case of client weight sharing and $N=\frac{pq}{K}$ in the case of no client weight sharing with alternating epochs. This hyperbola divides the regions where one technique perfors better over the other and both the feasible regions are shown in the theoretical schematic cartoon in Figure 3.
%Effect of parameters -

%1. Dataset size
%2. Eta
%3. Activations from split layer
%4. Number of clients
%5. Model Size

%data per client = $P/K$
%\vspace{-8mm}

\begin{figure}[!htb]
\minipage{0.32\textwidth}
  \includegraphics[width=\linewidth]{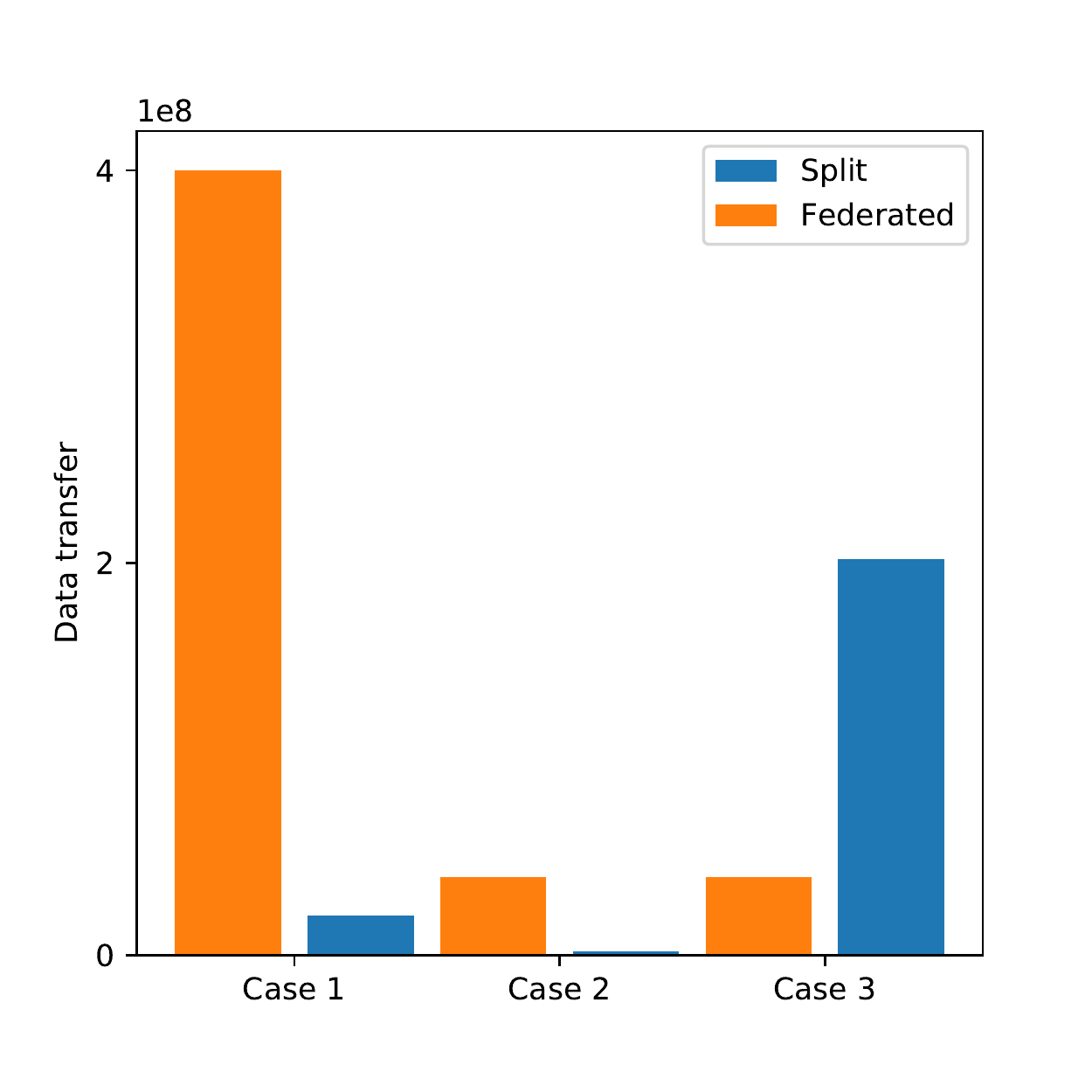}
 \caption{Use case corresponding to smart watches. Case-1 refers to a big model and high number of client setting. Case-2 refers to a relatively small model and Case-3 refers to even smaller model as well as low number of parameters.} \label{bar_client_a}
\endminipage\hfill
\minipage{0.32\textwidth}
  \includegraphics[width=\linewidth]{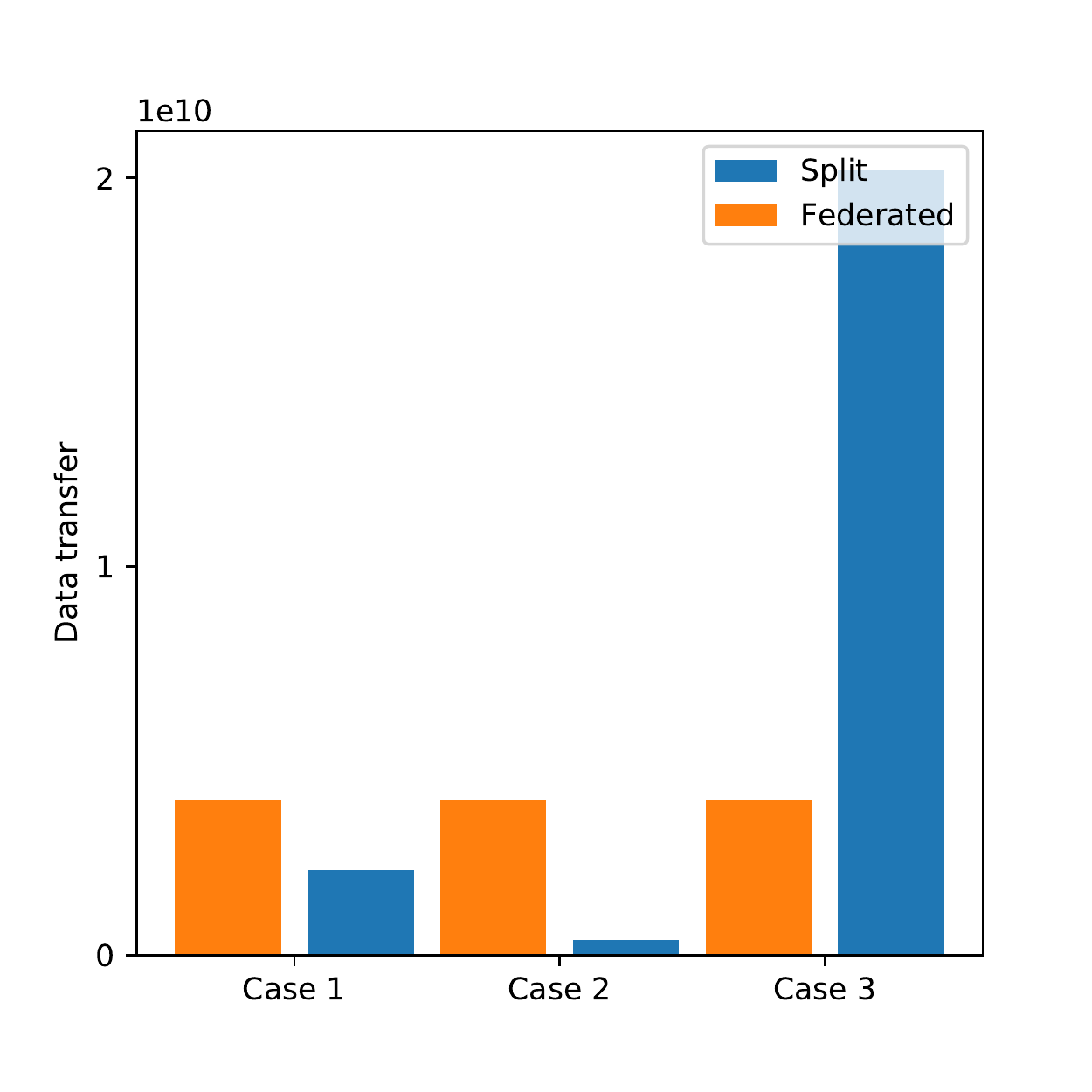}
 \caption{Use case for the healthcare scenario. Case-1 refers to a big model size and small number of clients. Case-2 has slightly higher number of clients and rest everything is same. Case-3 has bigger dataset and less number of clients.}\label{bar_client_b}
\endminipage\hfill
\minipage{0.32\textwidth}%
  \includegraphics[width=\linewidth]{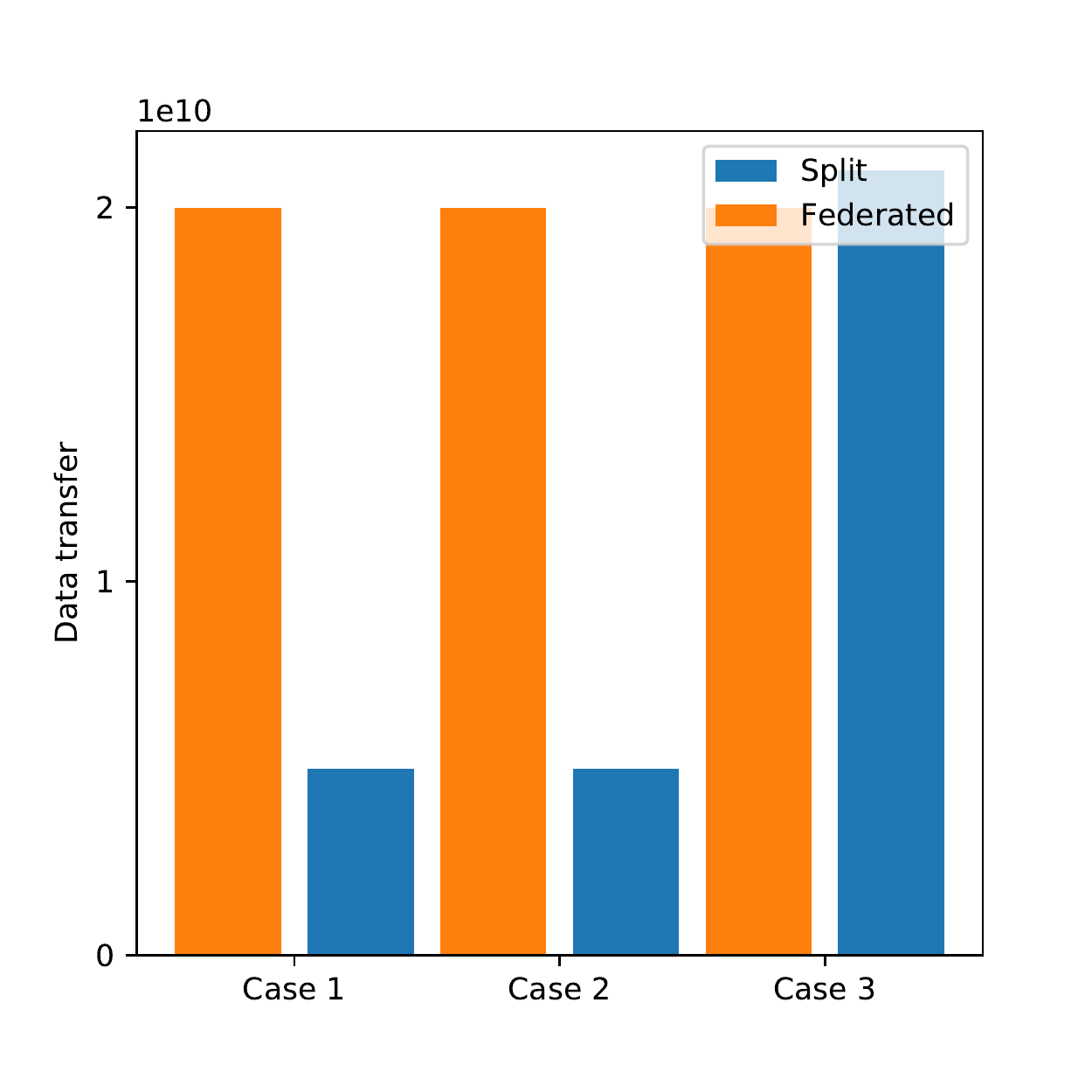}
  \caption{Amount of data transfer required for completing one round of training is shown for both split learning and federated learning like in adjacent figures and lower value on the y-axis means higher communication efficiency.}\label{bar_client_c}
\endminipage
\end{figure}

\section{Analysis}
\begin{figure}[!htbp]
\centering
\begin{subfigure}{0.31\textwidth}
    \label{curve_a}
    \includegraphics[width=4.6cm, height=5cm]{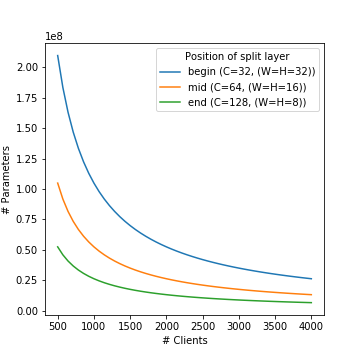}
    \caption{Number of clients in a setting where the number of parameters for the model is in the range of $10M-200M$.}
\end{subfigure}\hspace{12pt}%\hspace{0.2\textwidth}
\begin{subfigure}{0.31\textwidth}
    \label{curve_b}
    \includegraphics[width=4.6cm, height=5cm]{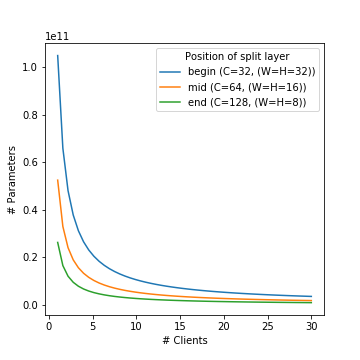}
    \caption{Small client setting where massive models can be fit like Transformers and AmoebaNet(~\cite{gpipe}).}
\end{subfigure}
\begin{subfigure}{0.31\textwidth}
    \label{curve_c}
    \includegraphics[width=4.6cm, height=5cm]{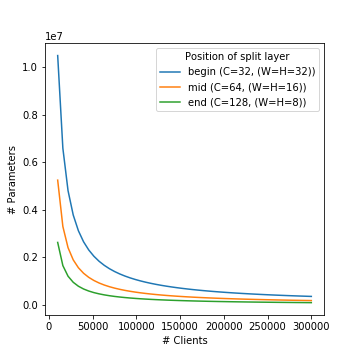}
    \caption{ Large client setting which allows a big range from tiny to large models to fit in is shown in this subfigure.}
\end{subfigure}
%\centering
\caption{Curve for comparing effeciency of Split learning and Federated learning setup. Similar to the figure~\ref{Split-learning}, upper half region shows the feasible region for relatively higher communication efficiency for Split Learning. The three curves in all three figures refer to the change in the position of split layer. We consider the size of commonly used activations in the CNN models during early layers.}
\label{comp_curves}
\end{figure}

We consider some of the real-life use cases which depict the relative efficacy of the two methods of split learning and federated learning.\\
First case is motivated by the scenario where the number of clients is in the range of millions. One concrete use case is smart watches (edge devices) which comprises of users in a diverse range from hundreds to millions. Figure(~\ref{bar_client_a}) presents all the three cases for the training under three different scenarios. Starting from the case-1, split learning setup has relatively lower data transfer due to the high number of model parameters and high number of clients. As we reduce the model parameters and number of clients in the case-3, the federated learning setup is relatively more communication efficient.
Second case, Figure(~\ref{bar_client_b}) is inspired from the healthcare setting where the clients are hospitals with large models but the number of clients is limited. In this case federated learning and split learning perform roughly same except the case three where federated performs better when dataset size is bigger and the number of clients is small.\\
Third case, Figure(~\ref{bar_client_c}) covers the use case for bigger institutions like biobanks where all the three parameters are in a high number. In case-1 and case-2, split learning setup outperforms the distributed learning because it scales well with the number of clients and the model parameters.\\
Figure(~\ref{comp_curves}) provides more practical scenarios of feasible curve of effeciency across a broad spectrum of parameters.

\section{Conclusion and future work} Our analysis suggests that the split learning setup becomes more communication efficient with increasing number of clients and is highly scalable with number of model parameters. Federated learning becomes more efficient with increasing the number of data samples especially when the number of clients is small or model size is small. In this work we also identify some of the use-cases where one would be more effective than the other in terms of communication efficiency. The analysis and discussion presented in this work would be benefecial for the distributed learning community to understand the potential benefits of both methods. This work could be extended by analyzing the resource utilization and number of epochs required for convergance of both distributed learning setups under different practical scenarios. \citep{split_1,split_2} shows that split learning converges drastically quicker than federated learning.% Since we compare communication efficiencies per epoch, the quicker converegence results in further higher communication efficiencies across all epochs needed to converge.

\bibliography{references}
\bibliographystyle{icml2019}
\end{document}